\documentclass{article}

\usepackage{arxiv}

\usepackage[utf8]{inputenc} 
\usepackage[T1]{fontenc}    
\usepackage{hyperref}       
\usepackage{url}            
\usepackage{booktabs}       
\usepackage{amsfonts}       
\usepackage{nicefrac}       
\usepackage{microtype}      
\usepackage{lipsum}		
\usepackage{graphicx}
\usepackage{natbib}
\usepackage{doi}
\usepackage{booktabs} 
\usepackage{subcaption}
\usepackage{graphicx}

\usepackage{xcolor} 
\usepackage{tikz}
\usetikzlibrary{shapes.geometric, arrows, positioning, automata,positioning,fit,backgrounds}
\tikzstyle{process} = [rectangle, minimum width=2cm, minimum height=1cm, text centered, draw=black, fill=white!30]
\tikzstyle{sum} = \tikzstyle{sum} = [draw, circle, minimum size=.5cm]
\tikzstyle{arrow} = [thick,->,>=stealth]

\usepackage{amsmath}

\usepackage{bm}
\usepackage{units}
\usepackage{float}
\usepackage{dblfloatfix}
\usepackage{algorithm}
\usepackage[noend]{algpseudocode}
\makeatletter
\def\BState{\State\hskip-\ALG@thistlm}
\makeatother

\title{The Effects of Learning in Morphologically Evolving Robot Systems}

\date{September 9, 1985}	
\date{} 					

\author{ \href{https://cs.vu.nl/ci/index.php/jie-luo/}{\includegraphics[scale=0.06]{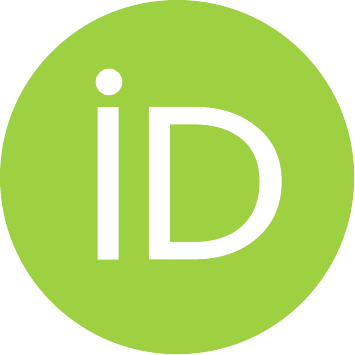}\hspace{1mm}Jie Luo}\\

	Department of Computer Science\\
	Vrije Universiteit Amsterdam\\
	\texttt{j2.luo@vu.nl} \\
	\And
	\href{https://}{\includegraphics[scale=0.06]{orcid.pdf}\hspace{1mm}Aart Stuurman} \\
	Department of Computer Science\\
	Vrije Universiteit Amsterdam\\
	\texttt{aart@astuurman.com} \\
	\And
	\href{https://jmtomczak.github.io/}{\includegraphics[scale=0.06]{orcid.pdf}\hspace{1mm}Jakub M. Tomczak} \\
	Department of Computer Science\\
	Vrije Universiteit Amsterdam\\
	\texttt{j.m.tomczak@vu.nl} \\
	\And
	\href{https://}{\includegraphics[scale=0.06]{orcid.pdf}\hspace{1mm}Jacintha Ellers} \\
	Faculty of Science\\
	Vrije Universiteit Amsterdam\\
	\texttt{j.ellers@vu.nl} \\
	\And
	\href{https://www.cs.vu.nl/~gusz/}{\includegraphics[scale=0.06]{orcid.pdf}\hspace{1mm}Agoston E. Eiben} \\
	Department of Computer Science\\
	Vrije Universiteit Amsterdam\\
	\texttt{a.e.eiben@vu.nl} \\
}

\renewcommand{\undertitle}{}

\hypersetup{
pdftitle={The Effects of Learning in Morphologically Evolving Robot Systems},
pdfsubject={Computing methodologies~Artificial intelligence, Computer systems ~Evolutionary robotics},
pdfauthor={Jie Luo, Jakub M. Tomczak, Agoston E. Eiben},
pdfkeywords={Evolutionary robotics,Embodied AI,Lifetime learning}
}

\begin{document}

\maketitle

\begin{abstract}
Simultaneously evolving morphologies (bodies) and controllers (brains) of robots can cause a mismatch between the inherited body and brain in the offspring. To mitigate this problem, the addition of an infant learning period by the so-called Triangle of Life framework has been proposed relatively long ago. However, an empirical assessment is still lacking to-date. In this paper we investigate the effects of such a learning mechanism from different perspectives. Using extensive simulations we show that learning can greatly increase task performance and reduce the number of generations required to reach a certain fitness level compared to the purely evolutionary approach. Furthermore, although learning only directly affects the controllers, we demonstrate that the evolved morphologies will be also different. This provides a quantitative demonstration that changes in the brain can induce changes in the body. Finally, we examine the concept of morphological intelligence quantified by the ability of a given body to learn. We observe that the learning delta, the performance difference between the inherited and the learned brain, is growing throughout the evolutionary process. This shows that evolution is producing robots with an increasing plasticity, that is, consecutive generations are becoming better and better learners which in turn makes them better and better at the given task. All in all, our results demonstrate that the Triangle of Life is not only a concept of theoretical interest, but a system architecture with practical benefits. 


\end{abstract}

\section{Introduction}
In this paper we investigate an evolutionary robot system where robot morphologies (bodies) and controllers (brains) are evolved simultaneously. Reproduction in such a system requires that a new body and a new brain are created and put together to form the robot offspring. In case of asexual reproduction one parent is sufficient and the new body and the new brain are obtained by random variation (mutation). For sexual reproduction (crossover) the bodies and the brains of the parents need to be recombined and the offspring is made up by the resulting body, driven by the resulting brain.

Regardless the specific way of reproduction, in general it cannot be assumed that the newly produced body and the newly produced brain form a good match. Even though the parents might have well-matching bodies and brains, recombination and mutation are stochastic operators and can shuffle the parental genomes such that the resulting body and brain do not form a good combination. Thus, the joint evolution of morphologies and controllers inherently leads to a potential body-brain mismatch. This problem has been originally noted by \citep{Eiben2013} and recently revisited in \citep{Eiben2020}. The proposed solution is the addition of learning. In terms of the robotic life cycle this means an extension of the straightforward two stages, the Morphogenesis Stage before 'birth' (when the robotic phenotype is constructed according to the instructions in the genotype) and the Operational Stage after 'birth' (when the robot is carrying out its task(s) and tries to reproduce). The third stage is the so-called Infancy Stage, following Morphogenesis, when the 'newborn' robot is learning to optimize the control of the inherited body and realize its maximum potential. The name Infancy is not only metaphorical, it also indicates an important detail: during this stage the robot is not fertile, that is, not eligible for reproduction. Algorithmically, this means that the fitness of a new robot is only calculated after the Infancy period, thus the robot can be subject to the (fitness based) mate selection procedure only after the learning is finished. 

This three-stage system architecture, dubbed The Triangle of Life, has been described in 2013, but to-date there are hardly any studies into the working of such systems. One reason is that in general there are not many papers on the joint evolution of morphology and controller, the majority of work in evolutionary robotics considers the evolution of brains within a fixed body. Furthermore, the computational costs can be prohibitive, or at least discouraging. To illustrate the severity of this issue let us note that the infant learning process is in fact a generate-and-test search process through the space of all possible controller configurations. Such a search algorithm will need to generate several new brains and test them all in the given body. This means several additional fitness evaluations\footnote{For the sake of simplicity, here we call all tests a fitness evaluation.} for each newborn robot. Giving newborn robots a learning budget of $N$ new controllers to be generated and tested during the Infancy stage implies that the total number of fitness evaluations will be $N$ times higher than evolving without learning. Of course, an appropriate value for $N$ depends on the system and the application at hand, but a few hundred or a few thousand learning trials seem common in the literature. 

The main goal of this research is to investigate the effects of learning in morphologically evolving robots. To this end, we set up a system where (simulated) modular robots can reproduce and create offspring that inherit the parents’ morphologies and controllers. Thus, in our system bodies and brains are both inheritable and therefore evolvable. Additionally, we make the brains not only evolvable, but also learnable. We achieve this by implementing a generic learning algorithm to optimize the inherited brain on any possible robot body. 

This system will be tested by giving it a problem to solve: developing robots for the task of targeted locomotion. This task will be interfaced to the evolutionary system by an appropriate fitness function that quantifies a robots ability to move towards a pre-specified target. Applying mate selection and survivor selection mechanisms that are based on this notion of fitness will drive evolution towards robots whose body and brain are well-suited for the given task.  

In order to investigate the effects of learning we will run the robot evolution process in two versions, one purely evolutionary and one where evolution and learning are combined. In the first one, the brain of a robot child is inherited from its parents and its fitness is established directly after 'birth'. In the second one, the inherited brain of a robot child is the starting point of a learning algorithm and the fitness of the robot will be based on its performance using the learned brain. 

Based on these experiments we address the issues of efficacy, efficiency, and the evolved morphologies. Regarding efficacy and efficiency, the main question is how the two algorithm variants compare in terms of the highest achieved performance and the time needed to achieve the best performance. Our underlying hypothesis is that in a real-world application learning forms a relatively fast and cheap way of reducing the number of slow and expensive evolutionary steps. To be specific, we can distinguish learning trials and evolutionary trials. An evolutionary trial equals to producing, activating and testing a new robot, including the manufacturing of its body, while a learning trial means producing, activating and testing a new controller in a given robot. Because controllers are digital entities (i.e. lines of code), learning trials require much less time, effort and material resources than evolutionary trials \citep{Eiben2021}. For instance, if the time of manufacturing a new robot is 12 hours, the time needed for computing and installing a new controller to be tried is 10 seconds and the duration of one fitness evaluation is one minute, then 1,000 evolutionary trials need approximately 501 days, while 1,000 learning trials cost less than one day. Our corresponding research questions are the following.

{\textbf{Research Question 1:}} How much 'savings' (in terms of evolutionary trials) can we achieve by adding infant learning as defined by the Triangle of Life architecture?

The answer to this question will indicate whether it is better to spend all allowable fitness evaluations on evolutionary trials (creation of a new robot consisting of a new body and a new brain) or should they be divided between evolutionary trials and learning trials (creation of a new brain in an existing body).

Concerning the morphologies, we will consider quantifiable morphological traits, e.g., size, symmetry, and compare the trends during the evolutionary process as well as the morphologies in the final populations.

{\textbf{Research Question 2:}} Will the addition of learning lead to different evolved morphologies?

Intuitively, one may expect no differences because learning is a process that only affects the brains, not the bodies. However, in the precursor of our current study we recently found that adding learning can lead to different morphologies \citep{Miras2020}.

Additionally, we investigate the evolution of {\it morphological intelligence}, defined by the learning potential of a given body. The idea is to wait with judgement about a newborn body and not judge it as-is right after birth with a semi-incidental controller, but to judge it as-it-can-be with a controller customized for that body through learning. Technically speaking we quantify morphological intelligence by what we call the {\it learning delta}, the performance difference between the inherited and the learned brain. Our definition and that of \citep{Gupta2021} are similar in spirit, but different in details. Specifically, \citep{Gupta2021} define morphological intelligence by the ability to learn
novel tasks, whereas the notion based on \citep{Miras2020} relates it to the given task.

{\textbf{Research Question 3:}} How does the morphological intelligence (measured by the learning delta) evolve over consecutive generations?


\section{Related Work}
In the field of Evolutionary Robotics the majority of studies consider the evolution of brains (controllers) within a fixed body (morphology). Research into systems where morphologies and controllers both evolve is scarce. This is not surprising, considering that the simultaneous evolution of morphologies and controllers implies two search spaces and the search space for the brain changes with every new robot body produced. This makes research more complex and challenging. 

\subsection{Evolvable morphology}
Similar research have been done recently with evolvable morphologies [\citep{Cheney2014}, \citep{Lipson2016}, \citep{Nygaard2017}, \citep{Cheney2018}, \citep{Nygaard2018}, \citep{Wang2019}, \citep{Liao2019}, \citep{DeCarlo2020}, \citep{Goff2021},  \citep{Gupta2021}, \citep{Medvet2021}]. Among all the studies, several approaches have been proposed to mitigate the body and brain mismatch effect on the population. 

Cheney \textit{et al.} \citep{Cheney2018} implemented a morphological innovation protection mechanism in which it allows additional optimization of the controller in a 'newborn' body by using mutation operator. During the protected Infancy period a robot can produce offspring, but it cannot be selected for removal from the population. Instead of protecting the population from being contaminated by the (possibly inferior) genes of a new individual, it protects a new individual from being wiped out by the (possibly superior) population members. 

Similarly, De Carlo \textit{et al.} \citep{DeCarlo2020} implemented protection in the form of speciation within their NEAT algorithm. The preservation of diversity in the population allowed new morphologies to survive, thus reducing the effects of body-brain mismatch. 

Nygaard \textit{et al.} \citep{Nygaard2017} demonstrated improvements in their ER system by introducing two phases during evolution. The first phase consists of both controller and morphology evolution, while during the second phase only the controller evolves in a fixed body. The results showed that, without the second phase, morphology and controller evolution led to sub-optimal controllers which required additional fine-tuning.

The work of Gupta \textit{et al.} \citep{Gupta2021} uses evolution for the robot bodies and reinforcement learning to optimize the controllers for the simulated robots. They introduced an asynchronous parallel evolution algorithm to reduce the computational budget by removing the typical generational aspect of an EA. 

In this paper, we use the Triangle of Life framework to integrate evolution and life time learning \citep{Eiben2013}. The essence is to have newborn robots perform a learning process that optimizes their inherited brain quickly after birth. An important additional feature is that newborn robots are considered to be infertile (i.e., not eligible for reproduction) until they successfully finish the learning period. This prevents that inferior genetic information is propagated and thus it saves resources. 

This method has been proven to be successful in Miras \textit{et al.} \citep{Miras2020}, however their research outcomes have a bias towards the snake like morphology and having gait learning as the fitness function is less practical.

\subsection{Controller learning algorithms}
With regards to the controller learning algorithms, several researches have been done recently by applying learning algorithms to the brains of robots with fixed bodies in order to produce optimal brains [\citep{Schembri2007}, \citep{Ruud2017}, \citep{Luck2019}, \citep{Schaff2019}, \citep{Jelisavcic2019}, \citep{Lan2020}, \citep{LeGoff2020}, \citep{Diggelen2021}].

In \citep{Diggelen2021}, three learning algorithms have been studied, namely: \textit{Evolutionary Strategies}, \textit{Bayesian Optimization} and \textit{Reversible Differential Evolution} (RevDE) \citep{Tomczak2020}. The study shows that the shape of the fitness landscape in Evolutionary strategies hints to a possible bias for morphologies with many joints. This could be an unwanted property for the implementation of lifetime learning because we want an algorithm that can work consistently on different kinds of morphologies. Bayesian Optimization is good at sample efficiency, however it required much more time comparing to the other two methods due to the higher time-complexity. Therefore RevDE outperforms among these three algorithms. 

However, RevDE generates three times as much new candidates and evaluating these comes with an extra computational cost while running the simulator. In this paper, we will use an advanced version of RevDE to alleviate this issue. This algorithm is introduced in the Algorithm section. Different from the researches mentioned above, we apply this learning algorithm not to the brains of fixed bodies but evolving morphologies. 

\section{Experiment Set-up}
The experiments have been carried out in Revolve (https://github.com/ci-group/revolve), a Gazebo-based simulator which enables us to test parts of the system as well as to set an entire environment for the complete evolutionary process. All experiments were performed using an infinite plane environment to avoid any extra complexity. We ran two experiments: experiment 1 works by running evolution alone. In this system, controllers are inheritable and the controller of the offspring is produced by applying crossover and mutation to the controllers of the parents. We refer to this experiment as Evolution Only throughout the paper. In experiment 2, controllers are not only evolvable, but also learnable. In these experiment, the controller of the offspring is produced by the learning algorithm that starts with the inherited brain. We refer to this experiment as Evolution + Learning throughout the paper.

\subsection{Robot morphology (Body)}
\subsubsection{Phenotype of body}
\begin{figure}[h]
\centering
  \includegraphics[width=100mm]{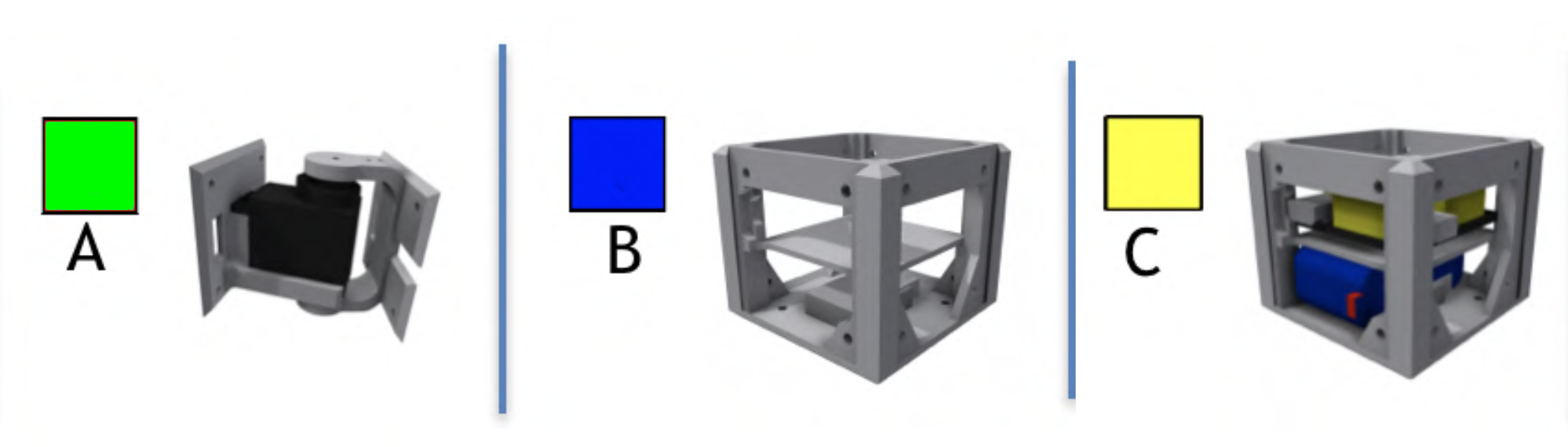}
  \caption{Modules of robots: The active hinge component (A) is a joint moved by a servomotor with attachment slots on both ends; The brick component (B) is a smaller cube with attachment slots on its lateral sides; The core component (C) is a larger brick component that holds a controller board and a battery. }
  \label{fig:components}
\end{figure}

The robots in Revolve are based on the RoboGen framework \citep{Auerbach2014}.
We use a subset of RoboGen's 3D-printable components: a morphology consists of one core component, one or more brick components, and one or more active hinges (see Figure \ref{fig:components}). The phenotype follows a tree-structure, with the core module being the root node from which further components branch out. Child modules can be rotated 90 degrees when connected to their parent, making 3D morphologies possible. The resulting bodies are suitable for both simulation and physical robots through 3D printing.

\subsubsection{Genotype of body}






The bodies are encoded in a Compositional Pattern Producing Network (CPPN) which was introduced by Stanley \citep{Stanley2007}. It has been demonstrated that with this encoding it is possible to evolve complex patterns for a variety of tasks [\citep{Clune2009}, \citep{Haasdijk2010}, \citep{Clune2011}, \citep{Auerbach2014a}, \citep{Goff2021}].

The structure of the CPPN has four inputs and five outputs. The first three inputs are the x, y, and z coordinates of a component, and the fourth input is the distance from that component to the core component in the three structure. The first three outputs are the probabilities of the modules to be a brick, a joint, or empty space, and the last two outputs are the probabilities of the module to be rotated 0 or 90 degrees. For both module type and rotation the output with the highest probability is always chosen; randomness is not involved.

The body's genotype to phenotype decoder operates as follows:\\
The core component is generated at the origin. We move outwards from the core component until there are no open sockets(breadth-first exploration), querying the CPPN network to determine the type and rotation of each module. Additionally, we stop when ten modules have been created. The coordinates of each module are integers; a module attached to the front of the core module will have coordinates (0,1,0). If a module would be placed on a location already occupied by a previous module, the module is simply not placed and the branch ends there.

\subsection{Robot controller (Brain)}
\subsubsection{Phenotype of brain}
We use Central Pattern Generators (CPGs)-based controllers to drive the modular robots. CPGs are biological neural circuits that produce rhythmic outputs in the absence of rhythmic input \citep{Bucher2015}. They are pairs of neurons ($x_i$,$y_i$) that drive rhythmic and stereotyped locomotion behaviors like walking, swimming, flying etc. in vertebrate species and they have been proven to perform well in modular robots \citep{Ijspeert2007}. 

\begin{figure}[hpt!]
\centering
  \includegraphics[width=85mm]{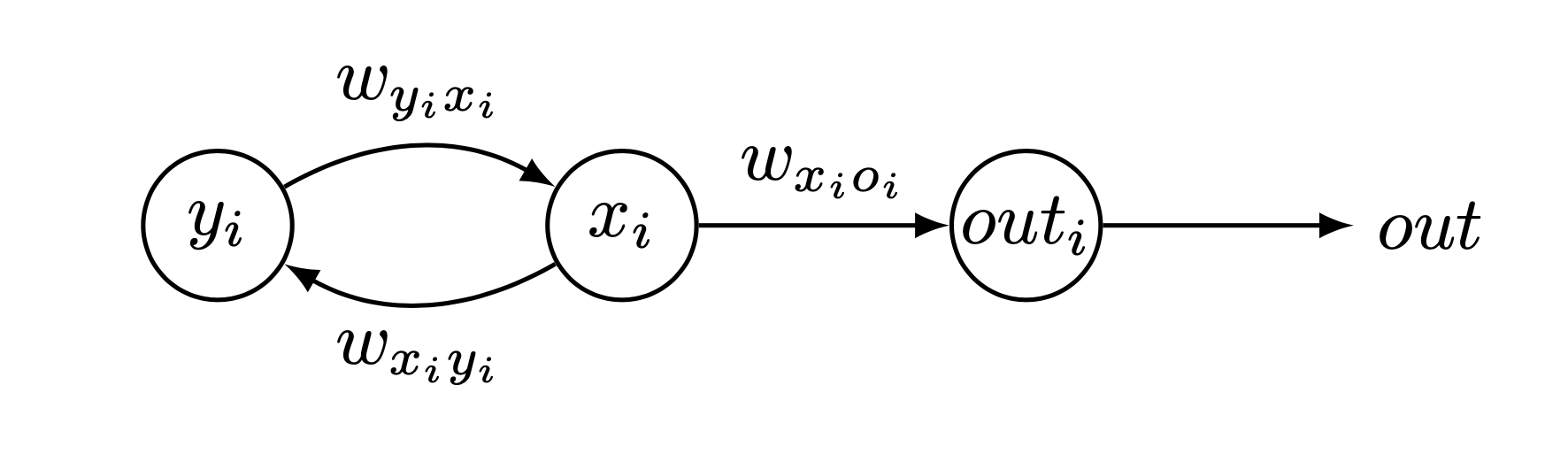}
  \caption{A single CPG. $i$ denotes the specific joint that is associated with this CPG. $w_{x_iy_i}$, $w_{y_ix_i}$, and $w_{x_io_i}$ denote the weights of the connections between the neurons, and out is the activation value of $out_i$-neuron that controls the servo in a joint.}
  \label{fig:CPG_single}
\end{figure}

In this study, the controllers are optimized for learning targeted locomotion. Each robot joint is associated with a CPG that is defined by three neurons: an $x_i$-neuron, a $y_i$-neuron, and an $out_i$-neuron that are recursively connected as shown in Figure \ref{fig:CPG_single}.

The change of the $x_i$ and $y_i$ neurons' states with respect to time is calculated by multiplying the activation value of the opposite neuron with a weight. To reduce the search space, we define $w_{x_iy_i}$ to be $-w_{y_ix_i}$ and call their absolute value $w_i$. The resulting activations of neurons $x$ and $y$ are periodic and bounded. The initial states of all $x$ and $y$ neurons are set to $\frac{\sqrt{2}}{2}$ because this leads to a sine wave with amplitude 1, which matches the limited rotating angle of the joints.

\begin{align}
\label{eq:single_CPG}
\begin{split}
	&\dot{x_i}  = w_{i}y_i \\
	&\dot{y_i}  = -w_{i}x_i
\end{split}
\end{align}

To enable more complex output patterns, CPG connections between neighbouring joints are implemented. Two joints are said to be neigbours if their manhattan distance is less than or equal to two. $x$ neurons depend on neighbouring $x$ neurons in the same way as they depend on their $y$ partner. Let $i$ be the number of the joint, $\mathcal{N}_i$ the set of indices of joints neighbouring joint $i$, and $w_{ij}$ the weight between $x_i$ and $x_j$. Again, $w_{ji}$ is set to be $-w_{ij}$. The extended system of differential equations is then:

\begin{align}\label{eq:ODE_CPG}
     \begin{split}
        &\Dot{x}_i = w_iy_i + \sum_{j \in \mathcal{N}_i} w_{x_jx_i}x_{j} \\
        &\Dot{y}_i = w_ix_i
     \end{split}
\end{align}

Because of this addition, $x$ neurons are no longer bounded between $[-1,1]$. To achieve this binding, we use a variant of the sigmoid function, the hyperbolic tangent function (tanh), as the activation function of $out_i$-neurons.

\begin{equation}
    out_{(i,t)}(x_{(i,t)}) = \frac{2}{1 + e^{-2 x_{(i,t)}}} - 1
    \label{eq:output}
\end{equation}

\subsubsection{Genotype of brain}
The structure of the brain largely depends on the body; only the weights between neurons can vary. Similarly to the morphology, we use CPPN as the genetic representation for the robot brain.

As each CPG corresponds to a joint, they have an associated three-dimensional position. The CPPN has six inputs: the coordinates of a CPG and the coordinates of another CPG connected to it. The output of the network is the weight between these CPGs. When querying for the weight between an $x$ and a $y$ neuron, the two input positions are both set to the same CPG coordinates. Although there are two weights between each two connected neurons, these are set to be the negative of each other. Because of this, the CPPN is only evaluated once, using coordinates of the neuron with the lowest index as the first three inputs.

\subsubsection{Controller steering}
In order to steer the modular robots, an additional steering policy is introduced. When the robot needs to turn right, joints on the right are slowed down, and visa versa. This does not lead to the correct steering behaviour for every robot, but we expect that robots will emerge that can use this policy successfully.

The magnitude of slow down, $g(\theta)$, is derived from $\theta$, the error angle between the target direction and the current direction. $\theta < 0$ means that the target is on the left and $\theta > 0$ means that the target is on the right. The target direction is calculated using the current absolute coordinates of the robot and the coordinates of the target point.
\begin{equation}
   g(\theta)= \left( \frac{\pi-\lvert\theta\rvert}{\pi} \right) ^{n}
    \label{eq:slow down factor}
\end{equation}

$n$ is a parameter that determines how strongly the joints slow down. In this experiment we choose $n=7$ based on manual fine-tuning.

Joints on the left side of the robot are controlled using the following formula:

\begin{equation}
signal = 
  \left\{
   \begin{aligned}
   &g(\theta) \cdot out & if\:\theta<0\\
   &out & if\:\theta\geq0 \\
   \end{aligned}
   \right.
   \label{eq:singal1}
\end{equation}

Analogously, for joints on right side:

\begin{equation}
signal = 
  \left\{
   \begin{aligned}
   &out & if\:\theta<0\\
   &g(\theta) \cdot out & if\:\theta\geq0 \\
   \end{aligned}
   \right.
   \label{eq:singal2}
\end{equation}

See Equation \ref{eq:output} for the meaning of 'out'.

\begin{figure}[hpt!]
\centering
  \includegraphics[width=410pt]{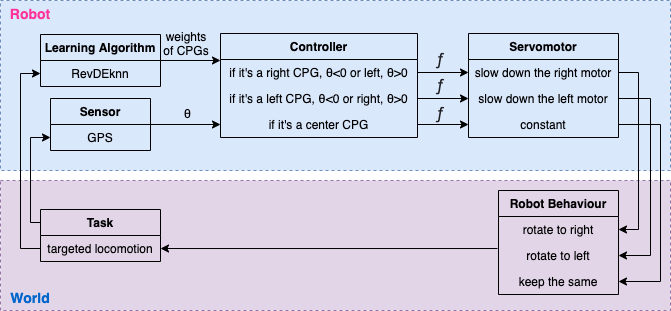}
  \caption{The overall architecture of how steering affects the joints. Error angle $\theta$ is calculated using the coordinates of the robot, its current heading, and its target. Function $f$ (Eq. \ref{eq:singal1} \& \ref{eq:singal2}), which uses the output of the CPGs and $\theta$, is then used to calculate the final signal going to the joints.}
  \label{fig:controller}
\end{figure}

Figure \ref{fig:controller} shows an overview of the complete control architecture.
\subsection{Algorithm}
It has been demonstrated that RevDE performs well to evolve controllers in modular robots for a given task \citep{Diggelen2021}.
However, it increases the computational cost of running the simulator by tripling the population. Here we introduce a surrogate model to overcome this issue \citep{Weglarz-Tomczak2021}. It uses the K-Nearest-Neighbor (K-NN) regressor to approximate the fitness values of the new candidates, then select the N most promising points. We refer to this approach as RevDEknn.

The learning algorithm works as follows:
\begin{enumerate}
    \item initialize a population with X samples. Here we add Gaussian noise to self-mutate the weights of CPGs ten times to obtain the initial population;
    \item assess all X samples;
    \item apply Reversible differential mutation operator and Uniform crossover operator:
    
Reversible differential mutation operator: Three new candidates are generated by randomly picking a triplet from the population, $(x_i,x_j,x_k)\in X$, then all three individuals are perturbed by adding a scaled difference in the following manner:
        \begin{equation}\label{eq:de3}
            \begin{split}
            y_1 &= x_i + F(x_j-x_k) \\
            y_2 &= x_j + F(x_k-y_1) \\
            y_3 &= x_k + F(y_1-y_2) 
            \end{split}
        \end{equation}
where $F\in R_+$ is the scaling factor. 

New candidates $y_1$ and $y_2$ could be further used to calculate perturbations using points outside the population. This approach does not follow a typical construction of an EA where only evaluated candidates are mutated. Further, we can express \eqref{eq:de3} as a linear transformation using matrix notation by introducing matrices as follows:

\begin{equation*}
    \left[\begin{array}{l}
y_{1} \\
y_{2} \\
y_{3}
\end{array}\right]=\underbrace{\left[\begin{array}{ccc}
1 & F & -F \\
-F & 1-F^{2} & F+F^{2} \\
F+F^{2} & -F+F^{2}+F^{3} & 1-2 F^{2}-F^{3}
\end{array}\right]}_{=\mathbf{R}}\left[\begin{array}{c}
x_{1} \\
x_{2} \\
x_{3}
\end{array}\right]
\end{equation*}

In order to obtain the matrix R, we need to plug $y_1$ into the second and third equation in (\ref{eq:de3}), and then $y_2$ into the last equation in (\ref{eq:de3}). As a result, we obtain N = 3X new candidate solutions and the linear transformation R is reversible.

Uniform crossover operator: the authors of \citep{Storn1997} proposed to sample a binary mask $m \in \{0, 1\}^D$ according to the Bernoulli distribution with probability p = P(md = 1) shared across D dimensions, and calculate the final candidate according to the following formula:
        \begin{equation}\label{eq:de2}
              v = m \odot y_n+(1-m) \odot x_n
        \end{equation}

Following general recommendations in literature \citep{Pedersen2010} to obtain stable exploration behaviour, the crossover probability p is fixed to a value of 0.9 and the scaling factor F is fixed to a value of 0.5;
\item apply K-NN to predict the assessment value of new samples (X+N) based on the K closest previously seen samples. The K-NN regression model is a non-parametric model that stores all previously seen individuals with their evaluations, and the prediction of a new candidate solution is an average over the K closest previously seen individuals (Table \ref{tab:parameters2}). In this paper, we set K = 3;
\item perform a selection over the population based on the prediction and select X samples;
\item repeat from step (2); 
\item terminate when the maximum iteration is reached. 

\end{enumerate}

\begin{figure*}[hpt!]
  \includegraphics[width=\linewidth]{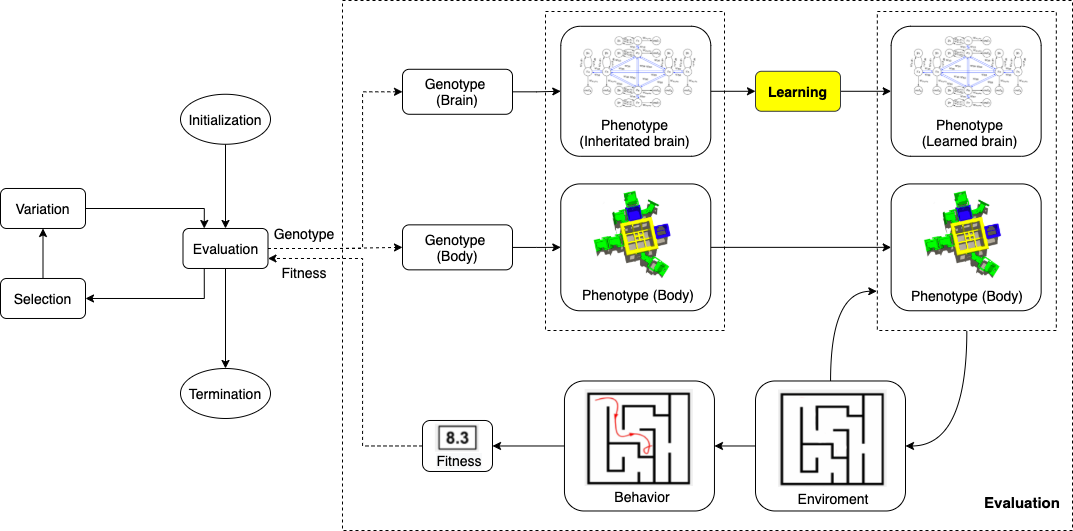}
  \caption{Evolution + Learning Framework: This is a general framework to create embodied robots via two interacting adaptive processes. An outer loop of evolution optimizes robot morphology via variation (mutation \& crossover) operations and an inner learning loop optimizes the parameters of a CPG controller. In the Evaluation box, we show examples of a robot morphology, controller, environment, behavior and fitness value.}
  \label{fig:E+L}
\end{figure*}

\begin{table}[ht!]
  \centering
 \caption{Parameters}\label{tab:parameters2}
    \begin{tabular}{l ll}
    \hline
    \textbf{RevDEknn}         & Value     & Description \\
    \hline
    $X$             & 10        & Initial sample size \\ 
    $F$             & 0.5       & Scaling factor\\ 
    $p$             & 0.9       & Crossover probability \\ 
    $k$             & 3         & Number of Nearest-Neighbors \\ 
    $g$             & 10        & Number of iterations \\
     \hline
    \end{tabular} 
\end{table}
In this paper, we apply RevDEknn to change the weights of the CPGs of modular robots to improve their controllers for targeted locomotion. For the outer loop, we use Evolutionary Algorithm (EA). The whole process is illustrated in Figure \ref{fig:E+L}. The code for carrying out the experiments is available online: \url{https://bit.ly/3t99vjC}. The pseudocode of combining EA and RevDEknn is shown below:


\begin{algorithm}[h!]
  \caption{Evolution + Learning}
  \label{alg:EA}
  \begin{algorithmic}[1]
    \State INITIALIZE robot population (genotypes + phenotypes with body and brain)  
    \State EVALUATE each robot  (evaluation delivers a fitness value)
    \While{not STOP-EVOLUTION}
        \State SELECT parents; (based on fitness)
        \State RECOMBINE+MUTATE parents' bodies; (genotype)
        \State RECOMBINE+MUTATE parents' brains; (genotype)
        \State CREATE offspring robot body; (phenotype)
        \State CREATE offspring robot brain; (phenotype)
        
        \State INITIALIZE brain(s) for the learning process;
        \While{not STOP-LEARNING}
            \State ASSESS offspring; (performance value)
            \State GENERATE new brain for offspring;
        \EndWhile 
        
        \State EVALUATE offspring with learned brain; (fitness value) 
        \State SELECT survivors / UPDATE population
          
    \EndWhile
 \end{algorithmic}
\end{algorithm}

\subsection{Fitness function}
CPG controllers that are learning targeted locomotion in modular robots with evolvable morphologies poses a black-box optimization problem. We therefore need to formulate a fitness function to reach our objective. In our system, the fitness function is not only used to evaluate the performance of robots but also serves as the guiding metric of learning controllers.

We define a fitness function for targeted locomotion that fulfills the following three objectives: 
\begin{enumerate}
    \item increasing the accuracy of travelling in not only one direction but all
    \item minimizing deviation with respect to the target points
    \item maximizing the speed with minimum length of the trajectory
\end{enumerate}

\begin{figure}[!ht]
\centering
\includegraphics[width=300pt]{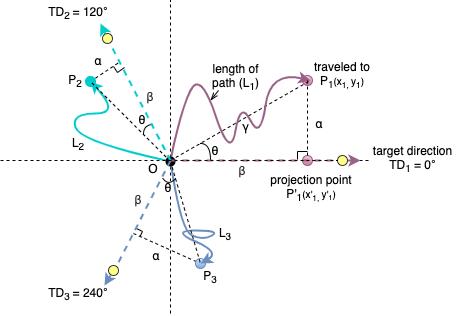}
\caption{\label{directionfit} Illustration of the fitness function. We use 3 targeted points (yellow dots) on 3 targeted directions $T_1=0$\textdegree (purple dotted line), $T_2=120$\textdegree (green dotted line) and $T_3= 240$\textdegree (blue dotted line). Point O is the starting position of the robot, $p_1(x_1,y_2)$ is an example of the end position of a robot learning target direction $T_1$. The purple line $L_1$ is the actual travel path between $(0,0)$ and $P_1$. The point $P'_1(x'_1,y'_1)$ is the projection of $P_1$ on the the target direction $T_1$. Same with the actual travel path $L_2$ and $L_3$ for $T_2$ and $T_3$ respectively.}
\end{figure}

The scenario of each robot's fitness evaluations with 3 targeted points along 3 targeted directions in our experiments is illustrated in Fig. \ref{directionfit}.

The elements we denote are:
\begin{enumerate}
\item $\alpha$, the distance between the projection point and the end point;
\item $\beta$, the distance between the projection point and the starting point $O(0, 0)$;
\item $\gamma$, the distance between the starting point and the end point;
\item $\theta$, the deviation angle between the actual direction of the robot locomotion and the target direction at time $T_{50}$;
\item $\mathcal{L}$, the path length of an individual's trajectory.
\end{enumerate}

The data we can collect from the simulator is as follows:
\begin{enumerate}
\item The coordinates of the core component of the robot at the start of the simulation (timestamp $T_0$) is approximate to O(0,0).
\item Every coordinate and the distance travelled (ds) per 0.2 sec incremental timestamps (dt). Thus allowing us to calculate the length of the path ($\mathcal{L}_1$, $\mathcal{L}_2$, $\mathcal{L}_3$) and plot the trajectory of the path.
\item The end points $P_1(x_1, y_1)$, $P_2(x_2, y_2)$, $P_3(x_3, y_3)$ are the coordinates of the core component of the robot at the end of the simulation (timestamp $T_50$) with 3 target directions.
\item The orientation of the robot in every 0.2 sec incremental timestamp plus the angle $\delta$ at timestamp $T_50$ as calculated by using the end coordinates.
\end{enumerate}

To achieve the first objective, we have to set multiple directions for training. In our fitness design, we select 3 fixed target directions equally distributed as $TD_1=0$\textdegree, $TD_2=120$\textdegree and $TD_3= 240$\textdegree. In this way, we avoid having robots whose  morphologies are suited for one direction only (which coincidentally matches the randomly selected direction). This presents the danger of these robots dominating the population, even though they don't perform well with the other directions.

To achieve the second objective, we have to minimize the angle $\theta$ which can be calculated as follows:
\begin{equation}
\theta = 
  \left\{
   \begin{aligned}
   &2 \pi - \left| \delta - TD \right|  \qquad \qquad \qquad (\left| \delta - TD \right| > \pi) \\
   &\left| \delta - TD \right|    \qquad \qquad \qquad \quad \quad ~(\left| \delta - TD \right| \leq \pi) \\
   \end{aligned}
   \right.
\end{equation}
Moreover, we also consider the deviation from the target direction by calculating the distance $\alpha$. By knowing the coordinates of the end position $P_n(x_n, y_n)$, we can use Euclidean distance to first calculate $\gamma$:
\begin{equation} 
    \gamma = \sqrt{(x_{n})^2 + (y_{n})^2}
    \label{eq:fitness_eq1}
\end{equation}

Then we can calculate $\alpha$ as follows: 

\begin{equation} 
    \alpha = \gamma \cdot sin(\theta)
    \label{eq:fitness_eq2}
\end{equation}

To have an efficient targeted locomotion, just moving in the right direction by achieving objective 1 and 2 is not enough, the robot should also move as fast as possible in the target direction which leads to the third objective. Given the same evaluation time, the higher speed means the longer distance travelled along the target direction. We calculate the distance travelled by the robot in the target direction by projecting the final position at timestamp $T_{50}$ onto the target direction, e.g. the point $P'_1(x'_1,y'_1)$ is the projection point of $P_1(x_1, y_1)$. 

We can calculate the projected distance in the target direction as follows: 

\begin{equation} 
    \beta = sign \cdot ( \gamma \cdot cos(\theta))
    \label{eq:fitness_eq3}
\end{equation}

where $\mathrm{sign} = 1$ if $\theta <\frac{\pi}{2}$ and $\mathrm{sign}=-1$ otherwise. This results in the $\beta$ being negative when the robot moves in the opposite direction.

Besides maximizing the projected distance, we also have to consider minimizing the length of the actual path of the robot's trajectory to have a robot that walks a the more efficient trajectory. We sum up all the ds values given the fixed evaluation time as the $\mathcal{L}$. 

In order to create a more robust fitness function, we introduce a penalty factor $\omega$ to fine-tune the result. $\omega$ is a constant scalar and in the experiments we set it to be 0.01.

The above amalgamates into our following fitness function:

\begin{equation} \tag{8}
    \mathcal{F}_{(\beta, \alpha, \theta, \mathcal{L})} = \frac{\left|\beta\right|}{\mathcal{L}+\varepsilon} \cdot \left(\frac{\beta}{\theta+1}-\omega \cdot \alpha\right)
\end{equation}

where $\varepsilon$ is an infinitesimal constant (set to $10^{-10}$ in the experiments). $\beta$ is divided by $\theta$ to further reward a small deviation angle (which is increased with $1$ to prevent from dividing by $0$). The function is maximised when $\beta$ is high, and $\mathcal{L}$, $\theta$, and $\alpha$ equal zero. In case of zero deviation from the target direction, $\mathcal{F}_{(\beta, \alpha, \theta, \mathcal{L})}=\beta$. 

To prevent passive individuals from taking over the population (potential local optimum), we set the fitness to zero for organisms that move less than 10 centimetres towards the target direction. When it comes to the implementation of the code, we use a distance of 10 meters from the original point (0,0) on the targeted directions to define the targeted points. 

\subsection{Experiment parameters}
In this paper we use a variant of the well-known ($\mu$ + $\lambda$) selection mechanism with $\mu =100$ and $\lambda = 50$ to update the population. An initial population of 100 robots is randomly generated as the first generation. In each generation 50 offspring are produced by selecting 50 pairs of parents through binary tournaments (with replacement) and creating one child per pair by crossover and mutation.  The next generation is formed by the top 50 parents plus the 50 offspring. The evolutionary process is terminated after 50 generations. In this research, the fitness value and performance value are the same. Therefore, for running Experiment Evolution Only, we perform (50+50×30) robots × 3 targeted directions = 4,650 fitness evaluations.


In Experiment Evolution + learning, for each evolutionary process, we applied RevDEknn on each robot 100 times (10 initial population multiplied by 10 generations). Using the RevDEknn algorithm the population tripled, meaning each robot morphology gets 300 controllers to learn targeted locomotion and the one with the highest fitness value will be the controller to match the morphology. This resulted in 100 assessments (300 divided by k=3 predictions) to simulate the robot’s limited field of view in the real-world. In total, $(50+50 \cdot 30) \cdot 3 \cdot 100=465,000$ fitness evaluations. 

We fix the evaluation time to be 50 seconds to balance computing time and accurately evaluating the task at hand.

To sum up, for running these 2 experiments, we perform 469,650 evaluations which amounts to $469,650 \times 50/60/60=6,523$ hours of (simulated) time. In practice, it takes about 13.59 days to run these experiments. All the experiments are repeated 10 times independently to get a robust assessment of the performance per data set. The experimental parameters we used in the experiments are described in Table \ref{tab:parameters}.

In order to have a better picture of the phylogenetic development and complete comparison with Evolution + Learning based on evaluations, we will keep on running Evolution Only experiment up till 3,099 generations (right now we have finished an average of 1,300 generations per run).

\begin{table}[ht!]
\caption{Main experiment parameters}
 \centering
\begin{tabular}{{p{0.25\linewidth} | p{0.1\linewidth}| p{0.5\linewidth}}}
\toprule
Parameters       & Value & Description                                    \\ \midrule
Population size  & ~100    & Number of individuals per generation     \\
Offspring size  & ~50    & Number of offspring produced per generation     \\
Mutation         & ~0.8   & Probability of mutation for individuals        \\ 
Crossover         & ~0.8   & Probability of crossover for individuals        \\ 
Generations      & ~30   & Termination condition for each run             \\ 
Learning trial  & ~100    & Number of the RevDEknn applied on each robot \\ 
Evaluation time  & ~50    & Duration of the test period per fitness evaluation in seconds \\ 
Tournament size  & ~2     & Number of individuals used in the parent selection - (k-tournament)		 \\ 
$\lambda \backslash \mu$ & ~0.5    & The ratio used in the survivor selection - ($\mu + \lambda)$  \\
Repetitions      &  ~10    & Number of repetitions per experiment \\ 
\bottomrule 
\end{tabular}
\label{tab:parameters}
\end{table}

\subsection{Performance measures}
To compare the two methods, we consider three performance indicators: \textit{efficacy}, \textit{efficiency} and \textit{morphological intelligence}. 

\subsubsection{Efficacy \& Efficiency}
We measure efficacy in two ways. Firstly, by the maximum fitness averaged over the 10 independent repetitions achieved at the end of the evolutionary process (30 generations). Since we consider targeted locomotion here, the quality is defined by the fitness value. As this measure can be sensitive to 'luck', we get more useful statistics by taking the average over 10 different runs. Secondly, another way to measure the quality of the solution is by given a same computational budget and measure which method find the best solution (highest fitness) faster.

Efficiency indicates how much effort is needed to reach a given quality threshold (fitness level). In this paper we use the number of Evaluations to Solution to measure it.

\subsubsection{Descriptors}
For quantitatively assessing morphological traits of the robots, we utilized the following set of morphological descriptors:\\
\textbf{Absolute Size}: Total number of modules of a robot body. It's a sum of all the structural bricks, hinges and one core-component with controller board. \\
\textbf{Proportion}: The length-width ratio of the rectangular envelope around the morphology. It is defined as following:
        \begin{align*}\label{eq:proportion}
              P =  \frac{p_s} {p_l}
        \end{align*}

where $p_s$ is the shortest side of the morphology, and $p_l$ is the longest side.\\
\textbf{Number of Bricks}: The number of structural bricks in the morphology.\\
\textbf{Relative Number of Limbs}: The number of extremities of a morphology relative to a practical limit. It is defined as following:
\begin{align*}
  L =
    \begin{cases}
      \frac{l}{l_{max}} & \text{if $l_{max} > 0$}\\
      0 & \text{otherwise}\\
    \end{cases}     
\end{align*}
\begin{align*}
  L_\text{max} =
    \begin{cases}
      \frac{2 \cdot (m-6)}{3} +(m-6)(mod 3) + 4 & \text{if $l_{max} > 0$}\\
      m-1 & \text{otherwise}\\
    \end{cases}   
\end{align*}

where m is the total number of modules in the morphology, l is the number of modules which have only one face attached to another module (except for the core-component) and $l_\text{max}$ is the maximum amount of modules with one face attached that a morphology with m modules could have, if containing the same amount of modules arranged in a different way.\\
\textbf{Symmetry}: The ratio between the horizontal and vertical symmetry given by the morphology.\\
\textbf{Branching}: The ratio between the total number of modules that have an attachment on all four possible lateral sides and the maximum number of modules that could have had an attachment on all sides given the morphology. 

We use the above six morphological descriptors to capture relevant robot morphological traits, and quantify the correlations between controller and morphology search spaces.

For quantitatively assessing behavioural characteristics of the robots, we utilized the displacement velocity descriptor which is defined as the distance between the last and first recorded position of the robot divided by the amount of travelling time, ignoring the path that was taken.
\section{Experiment Results}
\subsection{Efficacy \& Efficiency }
\begin{figure}[h]
  \includegraphics[width=\linewidth]{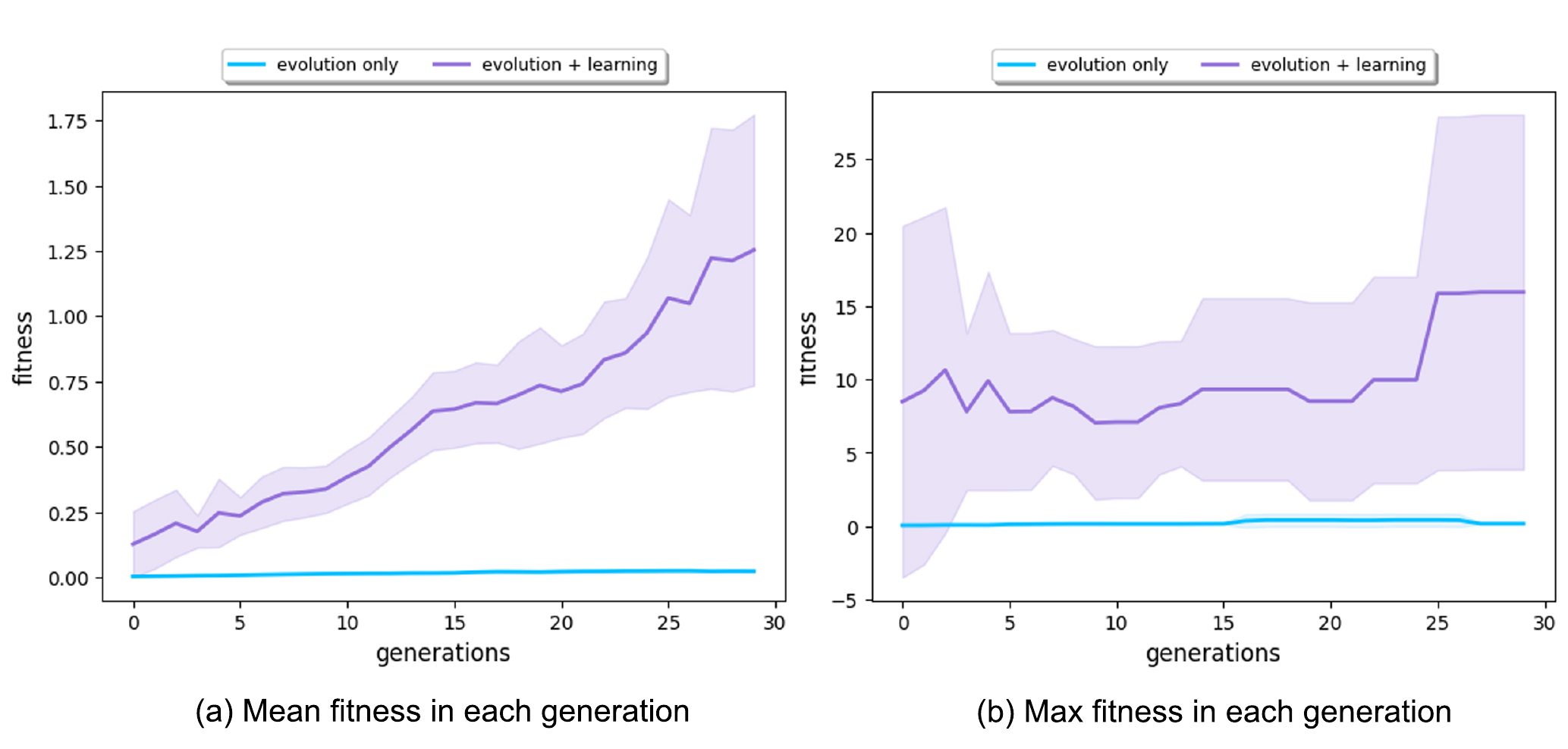}
  \caption{Subfigure (a) shows the mean fitness over 30 generations (averaged over 10 runs) for Evolution Only in blue and Evolution + Learning in purple. The Evolution + Learning method already surpassed in its first generation the fitness levels that the Evolution Only method achieved at the end of the evolutionary period. Subfigure (b) exhibits the maximum fitness in each generation (averaged over 10 runs). The shaded areas show the standard deviation. Note the different scales on the vertical axes.}
  \label{fig:fitness}
\end{figure}

Adding a life-time learning capacity to the system increased the fitness value of the robots, as depicted by Figure \ref{fig:fitness}. It shows that at the first generation, Evolution + Learning had already obtained an average fitness value that took Evolution Only the whole evolutionary period to achieve, i.e., 30 generations. In other words, Evolution + Learning at the first generation created only 100 robots and spent 30,000 evaluations while Evolution Only created 4,650 robots and 13,950 evaluations at generation 30. If we consider real physical robots, and assuming that the production cost of each robot (around 4 hours) is substantially higher than the computational budget (around 50 seconds), we can clearly see the advantage of introducing learning.
This was expected for two reasons: a) the number of evaluations performed by Evolution + Learning is around 100 times higher than with Evolution Only; b) in Evolution + Learning, robots have time to fine-tune their controllers to the morphologies they were born with. Instead, we are interested in verifying whether one method reaches the best solution faster than the other. 

\begin{figure}[h]
  \includegraphics[width=\linewidth]{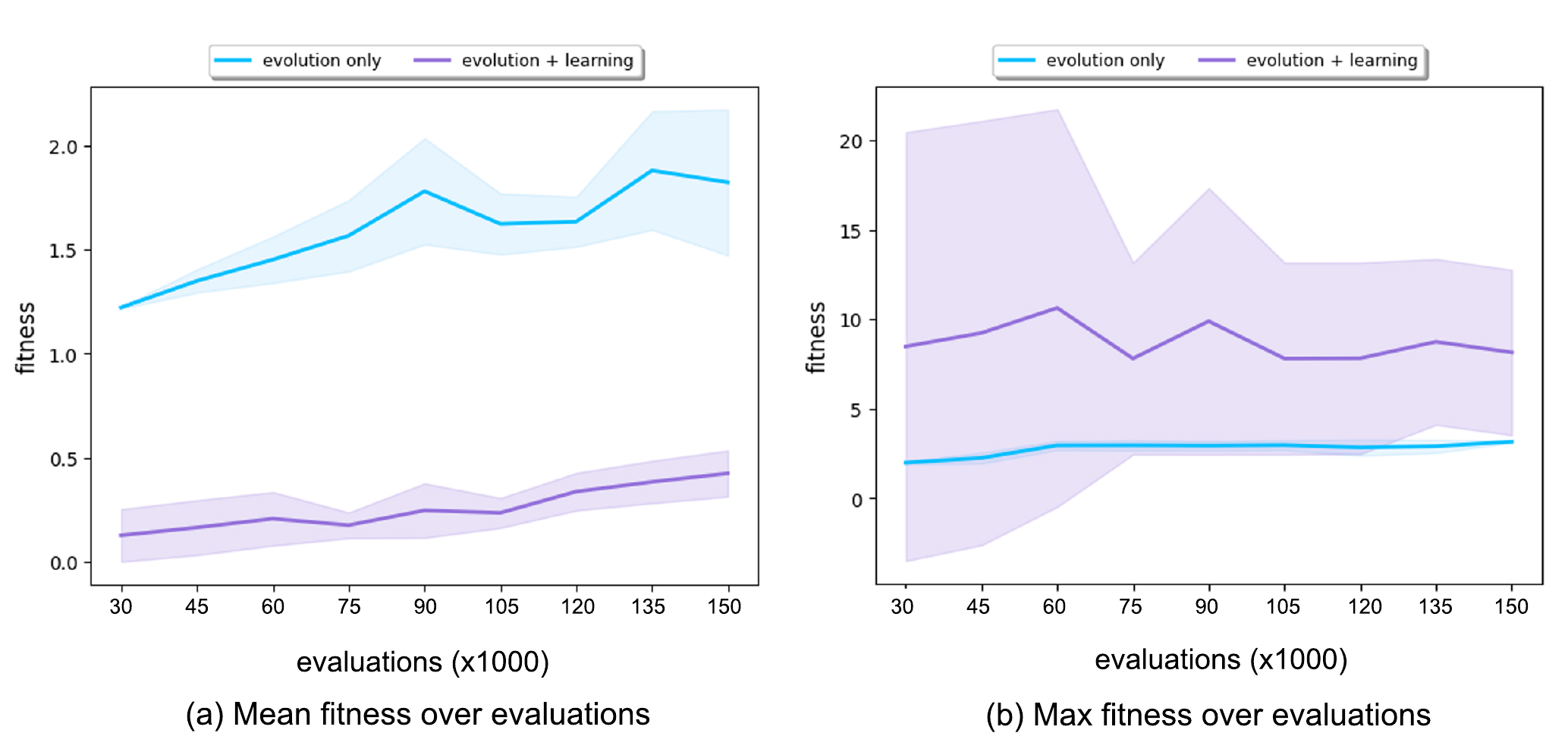}
  \caption{Subfigure (a) shows the mean fitness over 150,000 evaluations (averaged over 10 runs) for Evolution Only in blue and Evolution + Learning in purple. Subfigure (b) shows the maximum fitness over 150,000 evaluations (averaged over 10 runs) for Evolution Only in blue and Evolution + Learning in purple. The shaded areas show the standard deviation. Note the different scales on the vertical axes.}
  \label{fig:fitness-eval}
\end{figure}

Figure \ref{fig:fitness-eval} shows the mean and max fitness over the same number of evaluations. Evolution + Learning spent 150,000 evaluations created 500 robots at generation 9 while Evolution Only created 50,000 robots at generation 999. Very interestingly we can see that the average mean of Evolution Only is slightly higher than Evolution Learning while with regards to the max fitness, Evolution + Learning has much higher fitness and standard deviation. In other words, with learning, the robots with higher fitness and lower fitness differ more. This was unexpected, in order to have a better study of how the fitnesses of individual robots spread out in both methods, we plot the genealogical trees.

\begin{figure}[h]
  \includegraphics[width=\linewidth]{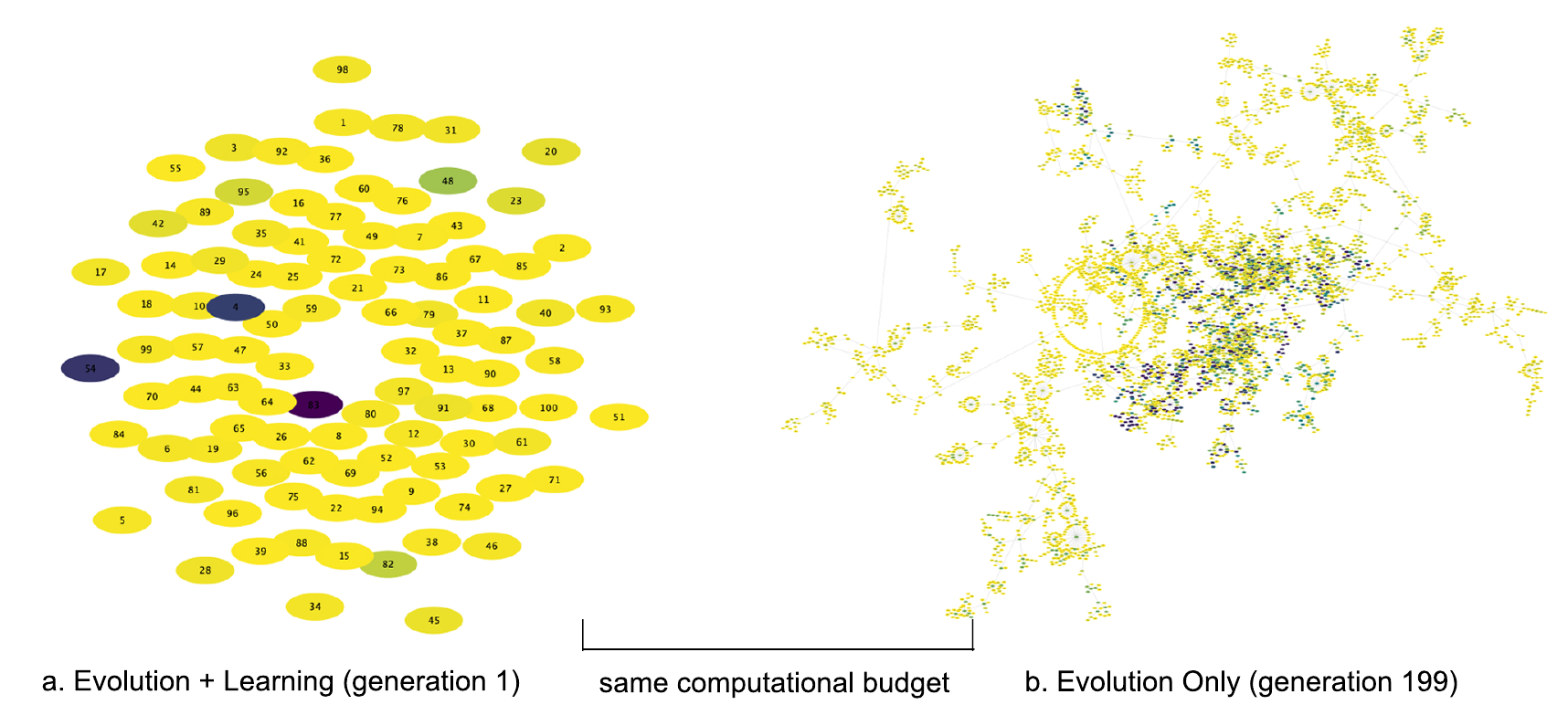}
  \caption{The genealogical trees of a single evolutionary run of both methods with the same computational budget: 30,000 evaluations. Each elliptical node represents a single robot. The node color indicates fitness (the darker the color the higher the fitness). The highest fitness found by Evolution + Learning is 31.90 (robot 83), whereas the highest fitness found by Evolution Only is 1.92 at the end of 199 generations.}
  \label{fig:phylogeny}
\end{figure}

Figure \ref{fig:phylogeny} shows the genealogical trees of a single evolutionary run of both methods with the same computational budget (evaluations = 30,000 and simulation time about 416.6 hours). In this computing time, Evolution Only evolved 199 generations while Evolution + Learning only generated the initial population of 100 without any offspring yet. In both plots, each elliptical node represents a single robot. The node color (from light yellow to dark purple) reflects fitness, with darker nodes indicating higher fitness. The plot on the left shows the first generation of Evolution + Learning without any connections to descendants yet. The plot on the right shows 199 generations of Evolution Only. The circle of nodes in the middle shows the initial population of 100. The nodes connected are their descendants. This tree demonstrates that founders with higher fitness intent to generate more lineages and higher fitness descendants. From both plots, we can see that although Evolution Only generates many fit robots, the best solution (max fitness: robot ID 83 with fitness of 31.90) is found in Evolution + Learning, whereas the max fitness found by Evolution Only is 1.92 at the end of 199 generations. In other words, if we want to achieve the target fitness level of 1.92, it takes Evolution Only 30,000 evaluations, while it only takes Evolution + Learning less than 249 evaluations (83 robots $\cdot$ 3 target directions).

\begin{figure}[h]
\centering
  \includegraphics[width=280pt]{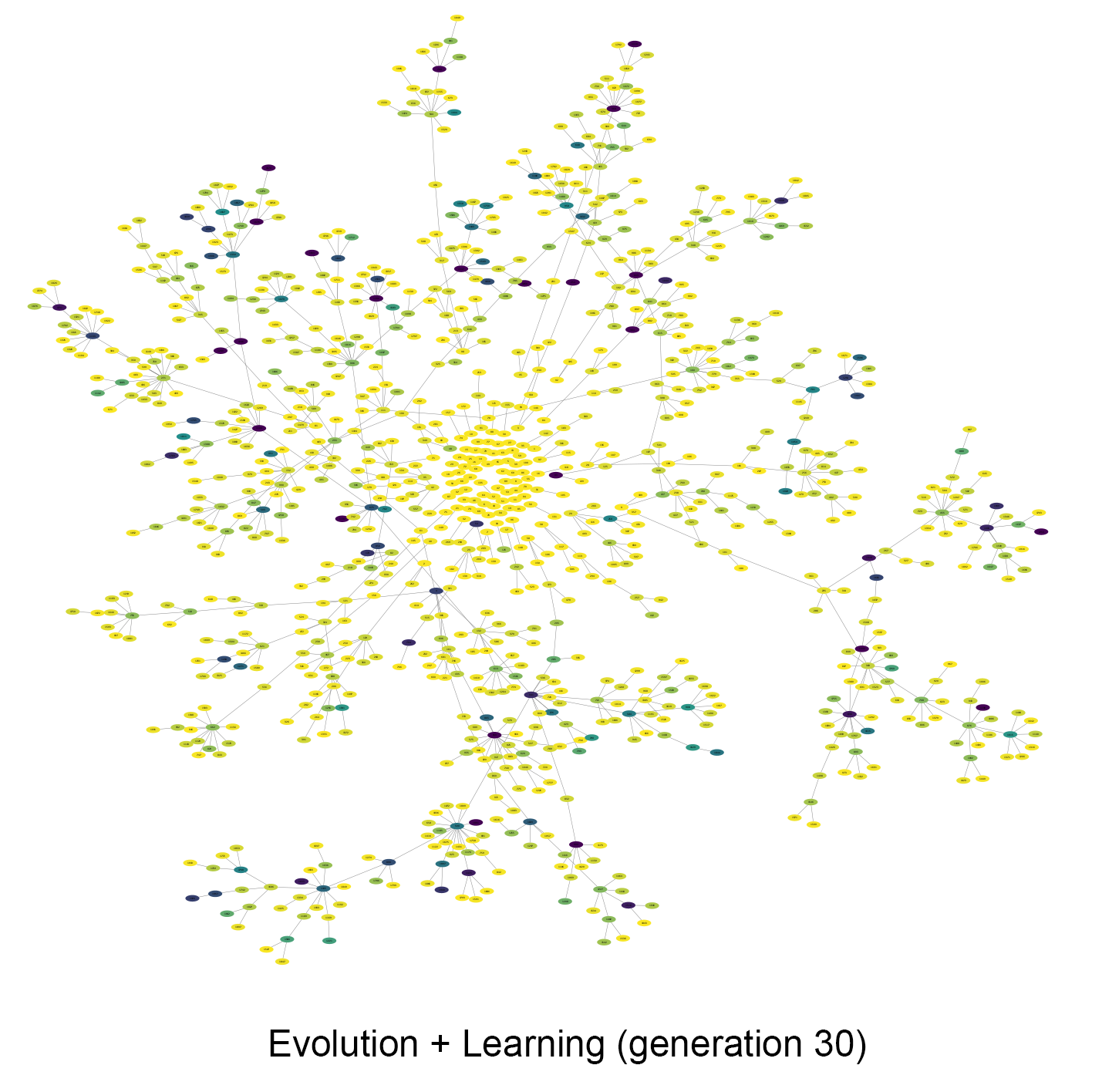}
  \caption{The genealogical tree of Evolution + Learning at generation 30. The descendants are spread out evenly across generations, as are the individuals with the highest fitness. It also demonstrates that descendants with high fitness can originate from ancestors with lower fitness (i.e. lighter nodes), rather than concentrating themselves in lineages that started with a high fitness in Evolution Only as shown in Figure \ref{fig:phylogeny}-b.('fitness begets fitness')}
  \label{fig:phylogeny2}
\end{figure}

Figure \ref{fig:phylogeny2} shows the genealogical tree of Evolution + Learning at the last generation. Similar to Figure \ref{fig:phylogeny}, the round circle in the middle shows the initial population of 100, then it's developed in a circular layout. The darker nodes indicating the higher fitness. Compare to Figure \ref{fig:phylogeny}-b, the descendants are more evenly spread out generation by generation, so do the max fitness individuals. It demonstrates that multiple lineages with descendants of high fitness can originate from founders with lower fitness (i.e. lighter nodes), rather than converging to one direction as shown in Figure \ref{fig:phylogeny}-b.

\subsection{Morphology}
\subsubsection{Morphological descriptors}
In \citep{Luo2021}, a study utilizing the same robot framework but using L-system as genetic encoding, a strong selection pressure for robots with few limbs was observed. In particular one single, long limb, i.e. a snake-like morphology. In this paper, we use CPPN as genetic encoding and the morphological traits of robots from Evolution Only are similar to \citep{Luo2021}, however we observe different morphological development trends in Evolution + Learning.

\begin{figure}[h]
  \includegraphics[width=\linewidth]{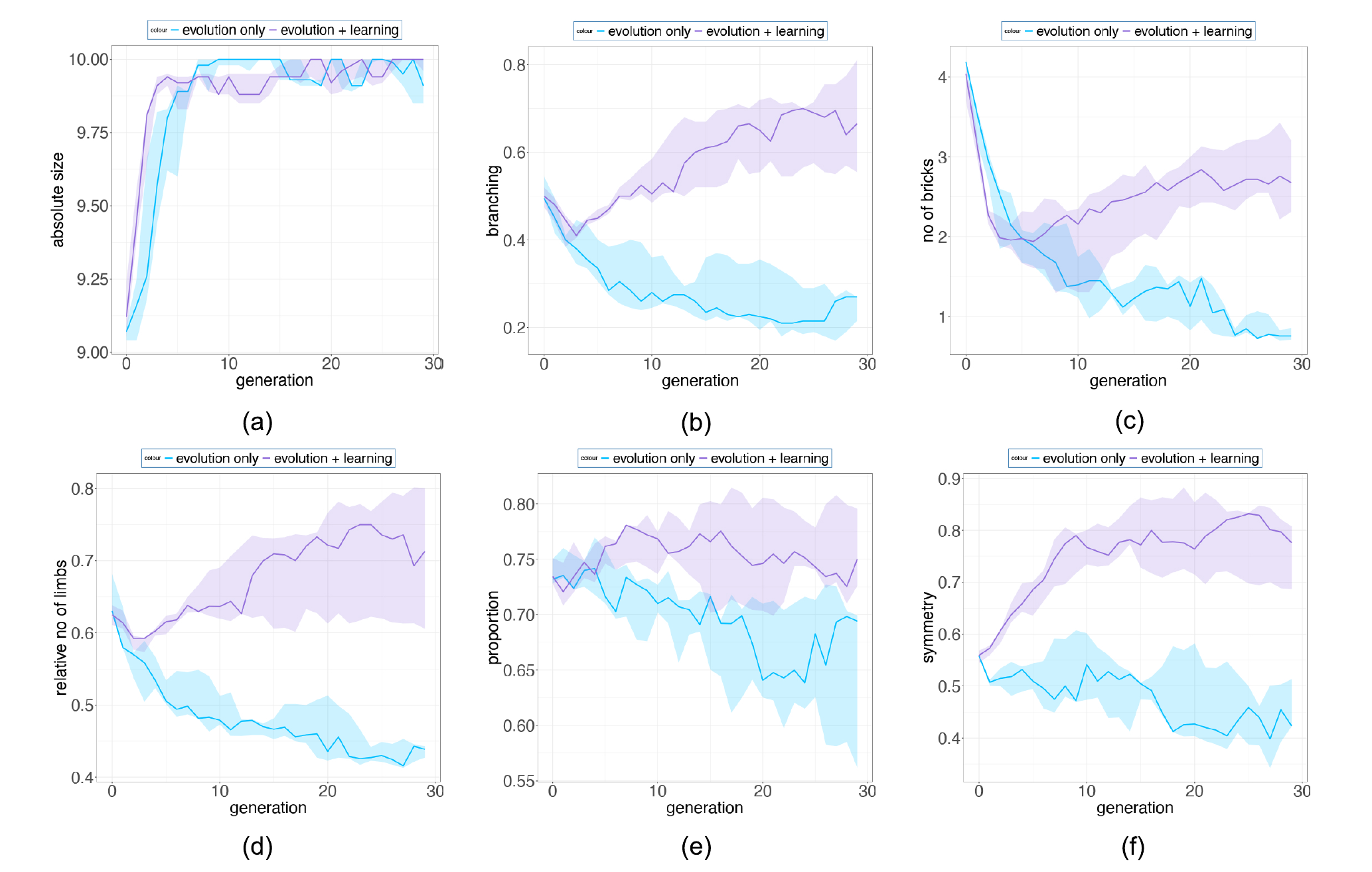}
  \caption{We selected 6 morphological traits that display a clear trend over generations and present the progression of their means averaged over 10 runs for the entire population. Shaded regions denote a 95\% bootstrapped confidence interval. (a), robots from both methods tend to develop bigger sizes over generations (b)(d), branching parameter and the relative no. of limbs from Evolution + Learning are increasing along the evolutionary timeline, however Evolution Only evolved in an opposite manner (c)(f), Evolution + learning robots tend to be more symmetric and have more bricks as they evolved compared to robots evolved in Evolution only (e), proportions from both methods show slightly decreasing trends, and the ratio of width and length of the robot morphology in Evolution + Learning is slightly higher over generations.}
  \label{fig:6 traits}
\end{figure}

We selected 6 morphological traits which display a clear trend over generations. In Figure \ref{fig:6 traits}, we see the progression of the mean of different morphological descriptors averaged over 10 runs for the entire population. 

\begin{figure}[h]
  \includegraphics[width=440pt]{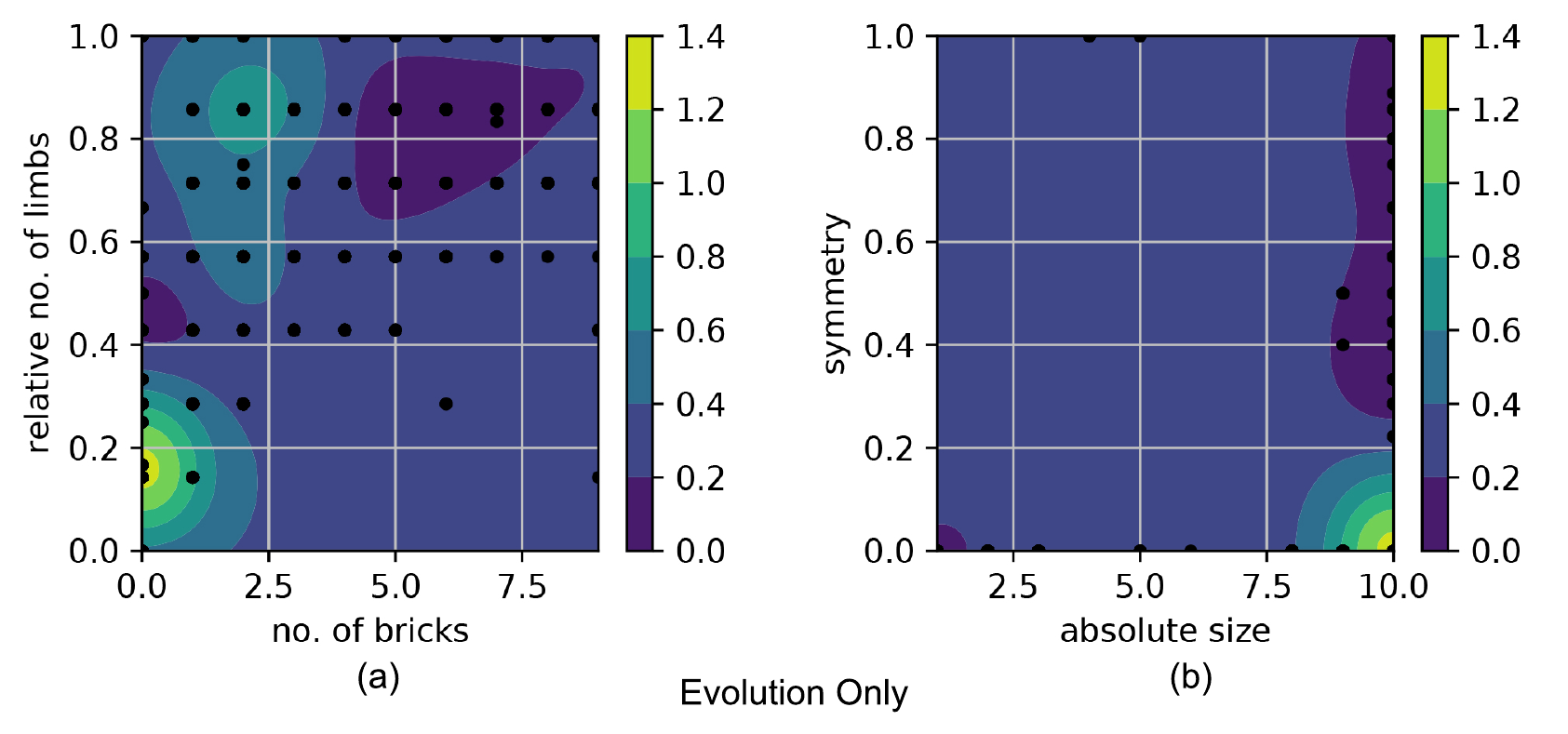}
  \centering
  \caption{Fitness landscape of Evolution Only using pairs of morphological measurements as coordinates. The color bars show the fitness level. (a) shows the fitness landscape of number of bricks over relative number of limbs, the robots with higher fitness values are in the morphological spaces of fewer bricks and limbs. (b) shows the fitness landscape of absolute size over symmetry. The robots with the highest fitness have a maximum absolute size of 10, however they are not symmetrical.}
  \label{fig:fitness_landscape_1}
\end{figure}

\begin{figure}[h]
  \includegraphics[width=440pt]{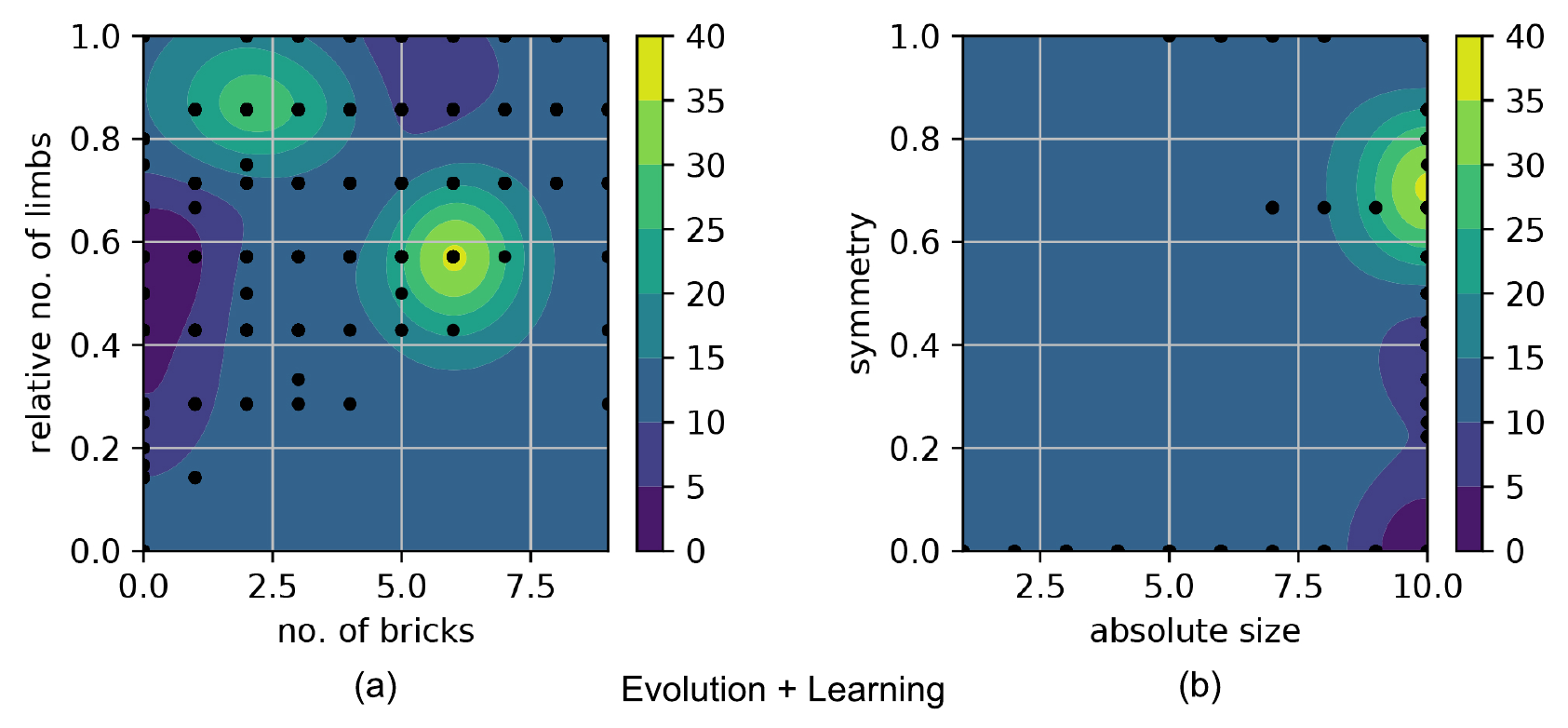}
  \centering
  \caption{Fitness landscape of Evolution + Learning. (a) the robots with higher fitness are in the regions where the contour color is brighter (yellow). (b) shows the fitness landscape of absolute size over symmetry. The highest fitness robots are highly symmetrical.}
  \label{fig:fitness_landscape_2}
\end{figure}

We present the fitness landscape plots (Figure \ref{fig:fitness_landscape_1} and Figure \ref{fig:fitness_landscape_2}) using pairs of morphological measurements as coordinates. Figure \ref{fig:fitness_landscape_1} - (a) and Figure \ref{fig:fitness_landscape_2} - (a) show the fitness landscape as a function of number of bricks over relative number of limbs of these two methods. In Evolution Only, the robots with higher fitness are in the morphological spaces of fewer bricks and limbs, while in Evolution + Learning, in the high fitness region where the contour color is brighter and wider are the robots with relatively more bricks and limbs. Figure \ref{fig:fitness_landscape_1} - (b) and Figure \ref{fig:fitness_landscape_2} - (b) show the fitness landscape of absolute size over symmetry of these two methods. In both methods, the robots with the highest fitness have a maximum absolute size of 10, however with regards to the symmetry, the highest fitness robots from Evolution + Learning are highly symmetrical whereas the ones from Evolution Only are not.

\begin{figure*}[hpt!]
    \centering
	\includegraphics[width=\linewidth]{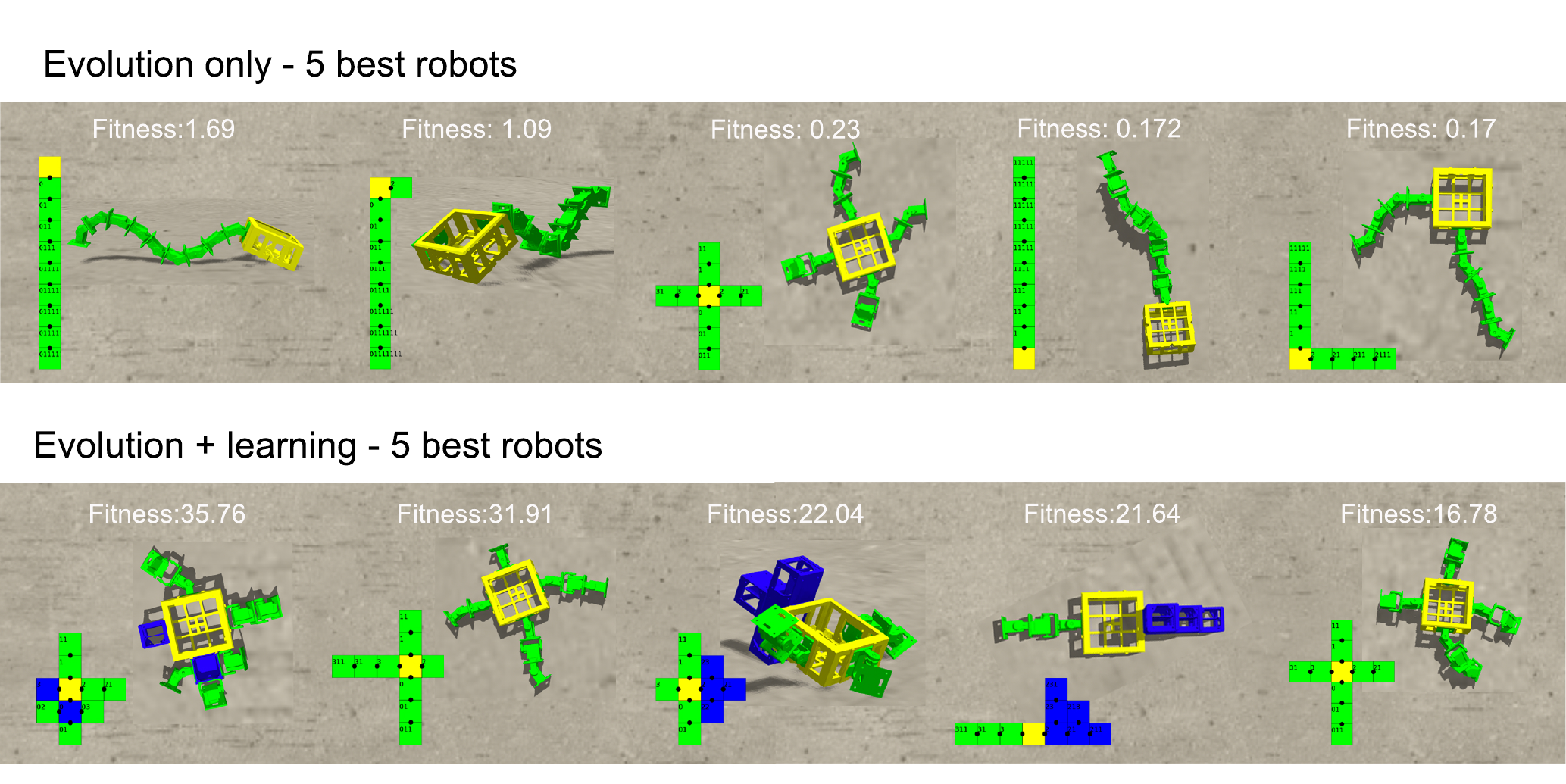}
	\caption {The 2D and 3D morphologies of 5 best robots from both methods. The best robots generated by Evolution+ Only are all converged into the morphology type that only contain hinges and no more bricks while the best robots from Evolution + Learning are more diverse. From the 3D morphologies we can see that the robots from Evolution Only are evolved with one layer while the robots from Evolution + Learning are developed in the horizontal dimension as well. With regards to the fitness, robots from Evolution Only are suffering from local optimum. In these 10 robots, we observe a same morphology from both methods (cross-shape), however the robot from Evolution + Learning has a much higher fitness value, once again it shows the importance of matching body and brain.}
	\label{fig:best robots}
\end{figure*}

In Figure \ref{fig:best robots}, it shows the 2D and 3D morphologies of 5 best robots from both methods. The best robots generated by Evolution+ Only are all converged into the morphology type that only contain hinges and no more bricks while the best robots from Evolution + Learning are more diverse.
From the 3D morphologies we can see that the robots from Evolution Only are evolved with one layer while the robots from Evolution + Learning are developed in the horizontal dimension as well. 
With regards to the fitness, robots from Evolution Only are suffering from local optimum. In these 10 robots, we observe a same morphology from both methods (cross-shape), however the robot from Evolution + Learning has a much higher fitness value, once again it shows the importance of matching body and brain. A video showing examples of robots from both types of experiments can be found in   https://youtu.be/4RJBpdNIR30.

\subsubsection{Morphological intelligence}
\begin{figure}[h]
\centering
  \includegraphics[width=300pt]{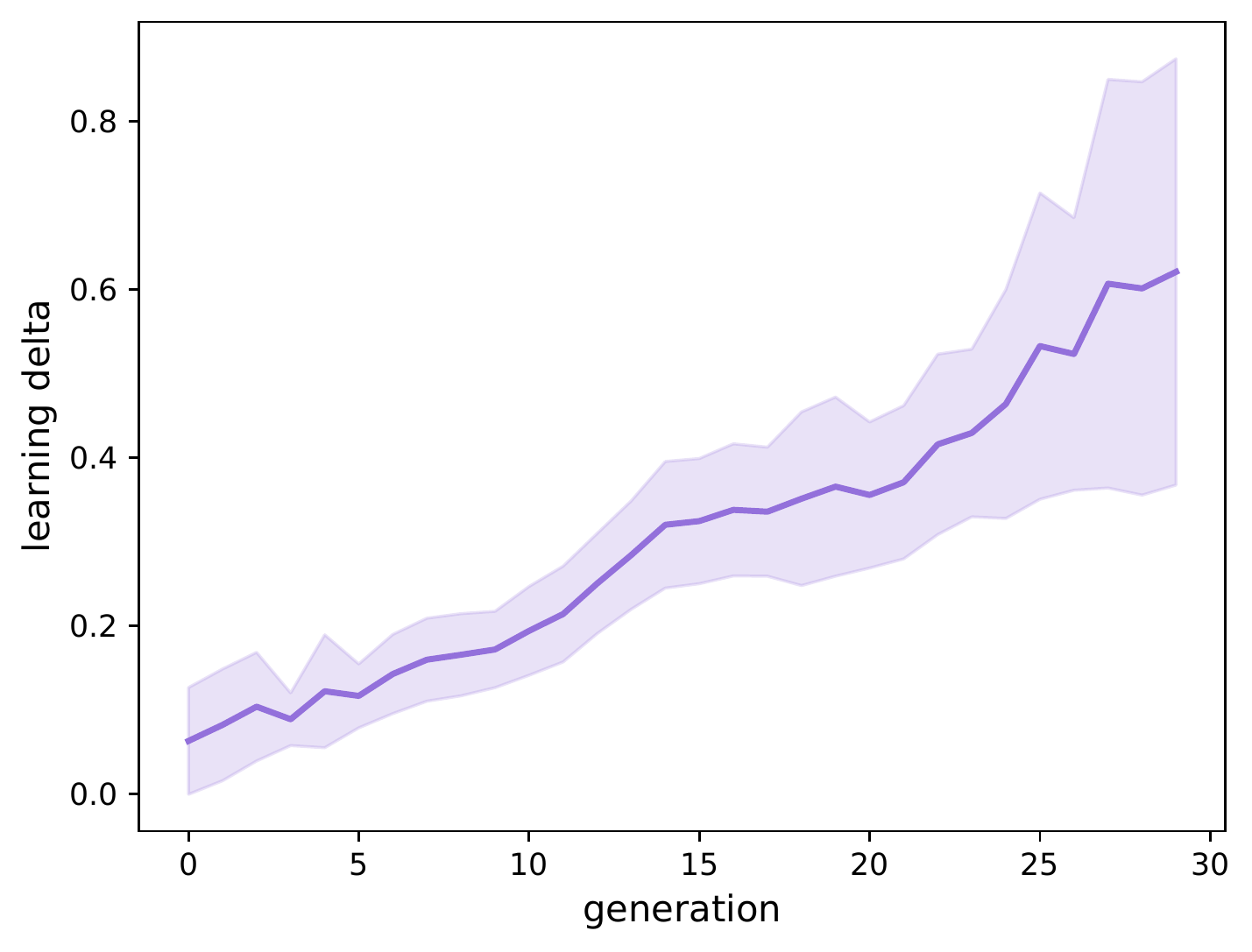}
  \caption{Learning $\Delta$: we quantify morphological intelligence by the learning delta, being the average fitness value after the parameters were learned minus average fitness value before the parameters were learned. Progression of the mean over 10 runs of the population. This growth is very steady. The observation indicates that the life-time learning led the evolutionary search to exploit the high performing morphological properties. In other words, the population turns into morphologies that are bigger and more symmetric which fit the brains better.}
  \label{fig:learning delta}
\end{figure}

In this paper, we consider a new concept: Morphological Intelligence. Morphology influences how the brain learns. Some bodies are more suitable for the brains to learn with than others. How well the brain learns can be empowered by a better body. Therefore we define the intelligence of a body as a measure of how well it facilitates the brain to learn and achieve tasks. In this paper, we quantify morphological intelligence by the learning delta, being the fitness value after the parameters were learned minus the fitness value before the parameters were learned. In Figure \ref{fig:learning delta}, we see that the average learning $\Delta$ of Evolution + Learning grows across the generations. 

\subsection{Behaviour}
\begin{figure*}[hpt!]
    \centering
	\includegraphics[width=\linewidth]{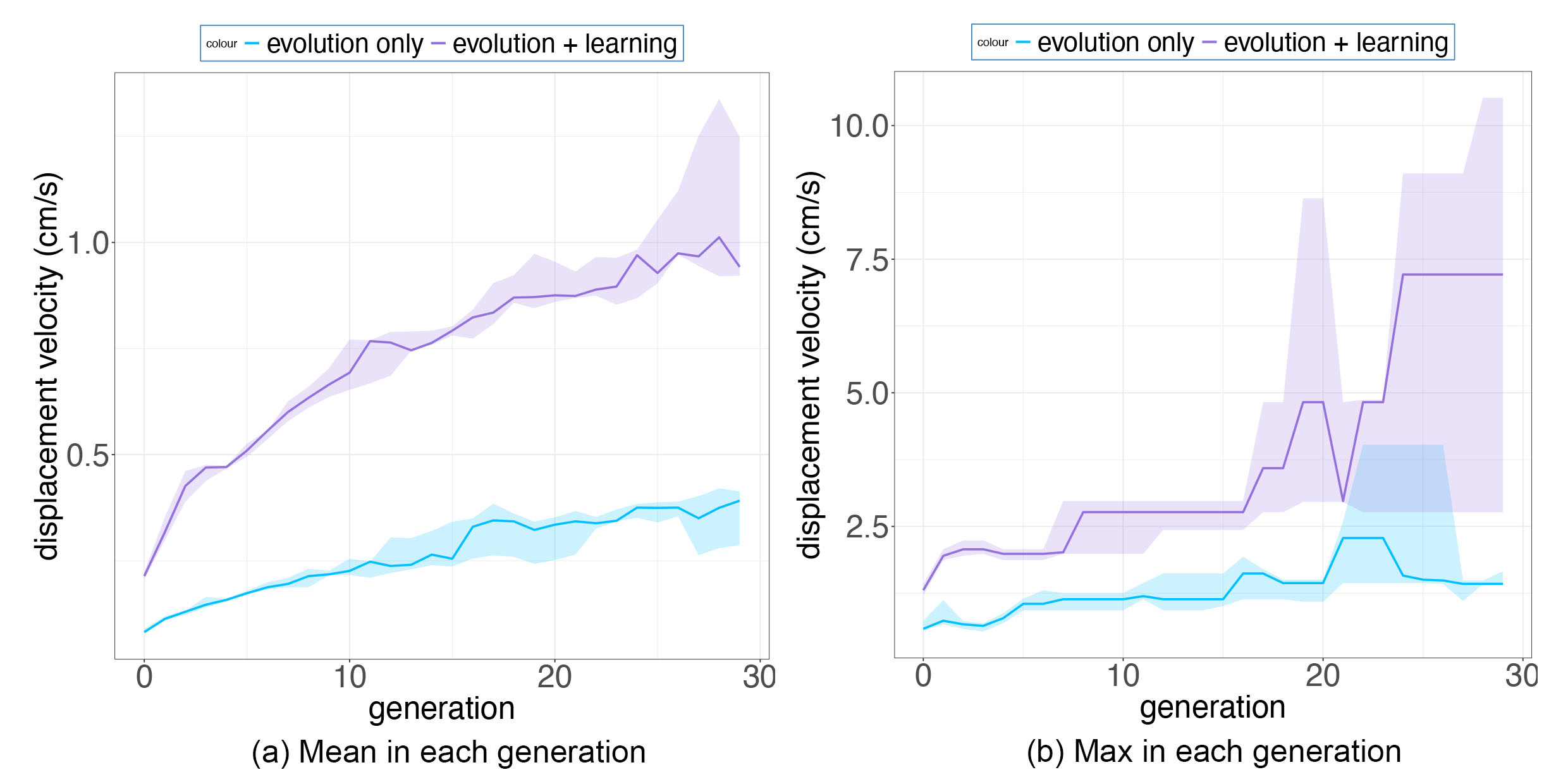}
	\caption {Subfigure (a) shows the mean displacement velocity over 30 generations (averaged over 10 runs) for Evolution Only in blue and Evolution + Learning in purple. Subfigure (b) exhibits the maximum displacement velocity in each generation (averaged over 10 runs). The shaded areas show the standard deviation. Note the different scales on the vertical axes.}
	\label{fig:verlocity}
\end{figure*}

\begin{figure*}[hpt!]
    \centering
	\includegraphics[width=\linewidth]{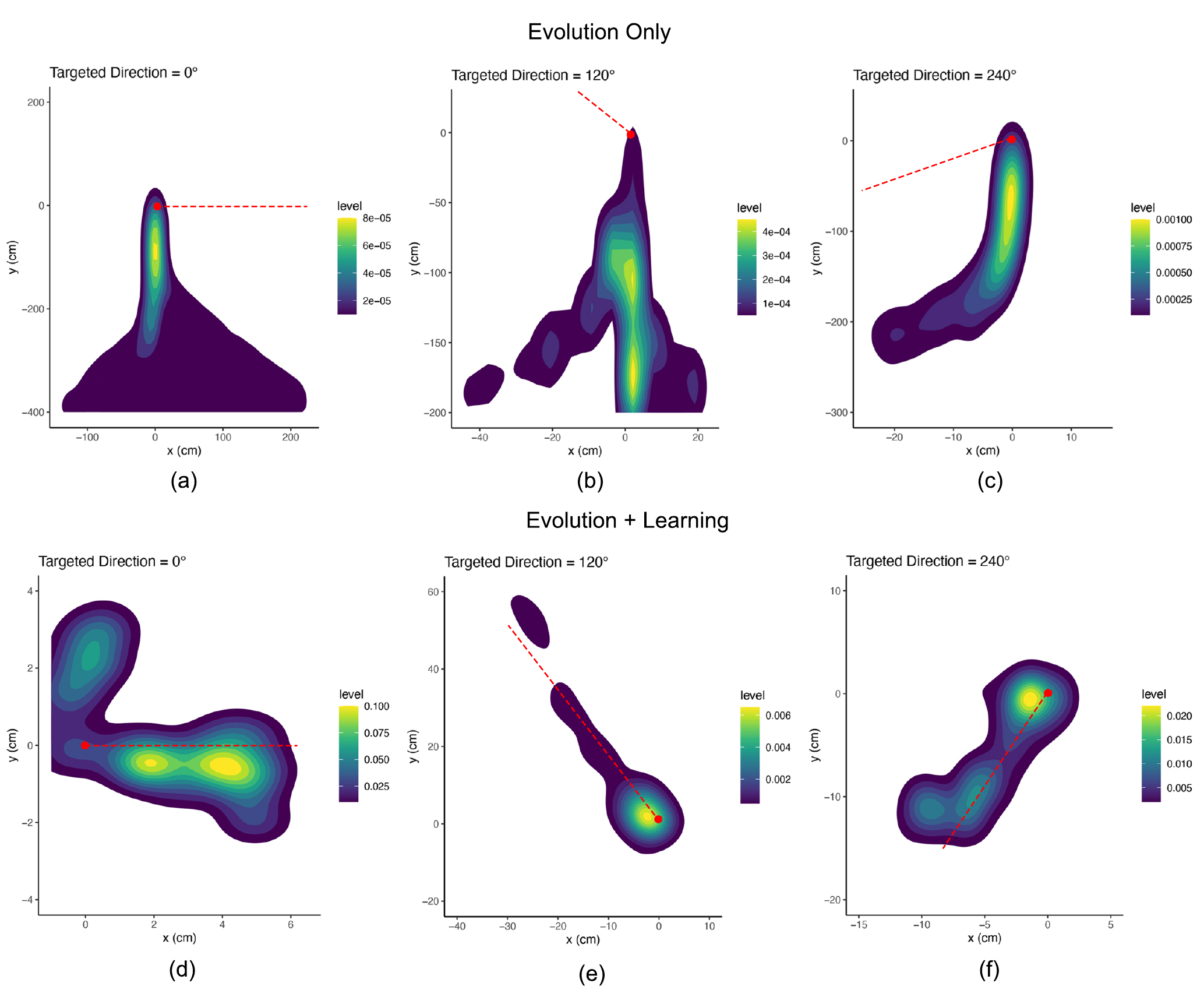}
	\caption {Trajectory density plots of the robots with maximum fitness in the last generation of all runs. The red dashed lines are the targeted directions. The red dots are the starting points.}
	\label{fig:trajectory}
\end{figure*}

Figure \ref{fig:verlocity} shows that both the mean and maximum displacement velocities of these two methods are significantly different. Especially in Figure \ref{fig:verlocity} - (b), the maximum displacement velocity in each generation from Evolution + Learning is much higher than Evolution Only, it can be due to the shorter path length, or the higher speed. To verify the reason, we inspect the trajectories of the robots. In order to have a picture of several robots' behavior rather than regarding individuals, we use a density plot to visualize the trajectories of the highest fitness robots in the last generation of all runs (Figure \ref{fig:trajectory}). From Figure \ref{fig:trajectory} - a, b, c, we observe that the best robots from Evolution Only in a given time travelled a long distance, indicating a high speed, however they are not efficiently moving towards the target directions. From Figure \ref{fig:trajectory} - d, e, f, we see that the best robots from Evolution + Learning travelled less distance compared to Evolution Only but very efficiently towards the targeted directions. 

\section{Conclusions and Future work}

The data on the efficacy measurements shows that Evolution + Learning can achieve much higher levels of fitness with the same number of fitness evaluations than Evolution Only. The differences are very large, speeds obtained by Evolution + Learning are three to four times higher than those through Evolution Only, cf. Figure \ref{fig:verlocity}. This renders the issue of efficiency quite irrelevant as it does not make much sense to compare the number of evaluations needed to achieve the same level of fitness. 

These outcomes provide an answer to our first research question clearly showing that spending time on infant learning can lead to massive savings in terms of evolutionary trials; Evolution + Learning achieved speed levels in eight generations that required 21 generations for Evolution Only. Thus, we can formulate a recommendation for users and experimenters: it is advisable to divide the number of allowable fitness evaluations between evolutionary trials and learning trials instead of spending them all on evolution only. The balance, i.e., the adequate number of learning trials per newborn robot, will differ by the specific numbers of the given application, but our results demonstrate that the Triangle of Life is not only a concept of theoretical interest, but a system architecture with practical benefits.  

Regarding the second research question, the data on the morphological traits proved that the learning approach leads to differently evolved morphologies. This is highly interesting if we realize that the difference between Evolution Only and Evolution + Learning is only affecting the controllers. In other words, our results show that modifying the way brains are treated results in pronounced changes to the bodies. Thus, we clearly demonstrate how the brains can shape the bodies through affecting task performance that in turn changes the fitness values that define selection probabilities during evolution. This is an interesting inversion of the classic statement that the bodies shape the brains \citep{Pfeifer2007}.

Our third research question concerns the evolution of morphological intelligence as quantified by the learning delta. The crucial plot to answer this question is Figure \ref{fig:learning delta} that shows how the delta is increasing over the course of evolution. This is a powerful demonstration of a how consecutive generations are becoming better and better learners which in turn makes them better and better at the given task. Putting it differently, evolution is producing robots with an increasing plasticity. Of course, we are not claiming that the learning delta values can grow without limits, but in the curve of the thirty generations we could afford with the current computing hardware there is no sign of a plateau.

For future work, we will study more tasks related to locomotion and object manipulation not only in isolation, but also in combination, such that multiple task-abilities determine the fitness together. Furthermore, we will investigate Lamarckian evolution, where the learned brain features are coded back to the genotypes, thus becoming inheritable. 

To this end we have to carefully consider a suitable genetic representation which allows us to partly invert the genotype-phenotype mapping of the brains. 



\clearpage
\bibliographystyle{unsrtnat}
\bibliography{references}  






\end{document}